\documentclass[conference]{IEEEtran}
\IEEEoverridecommandlockouts
\usepackage{amsmath,amssymb}
\usepackage{graphicx}
\usepackage{booktabs}
\usepackage{array}
\usepackage{multirow}
\usepackage{siunitx}
\usepackage{cite}           
\usepackage{xurl}  
\usepackage{soul}
\usepackage{xcolor}
\usepackage{xspace}
\usepackage[hidelinks]{hyperref}
\newcommand{\tripletgcn}{\textit{Triplet-GCN}\xspace}
\newcommand{\gcn}{\textit{GCN}\xspace}
\newcommand{\mlp}{\textit{MLP}\xspace}
\newcommand{\ehr}{\textit{EHR}\xspace}
\begin{document}
\onecolumn

\title{Sepsis Prediction Using Graph Convolutional Networks over Patient–Feature–Value Triplets}

\author{
\IEEEauthorblockN{
Bozhi Dan\textsuperscript{1},
Di Wu\textsuperscript{1,4}\textsuperscript{*},
Ji Xu\textsuperscript{2},
Xiang Liu\textsuperscript{3,4},
Yiziting Zhu\textsuperscript{3,4},
Xin Shu\textsuperscript{3,4},
Yujie Li\textsuperscript{3,4},
Bin Yi\textsuperscript{3,4}
}
\IEEEauthorblockA{\textsuperscript{1}\,College of Computer and Information Science, Southwest University, Chongqing 400715, China}
\IEEEauthorblockA{\textsuperscript{2}\,State Key Laboratory of Public Big Data, Guizhou University, Guiyang 550025, China.}
\IEEEauthorblockA{\textsuperscript{3}\,Department of Anesthesiology, Southwest Hospital, Third Military Medical University\\ (Army Medical University), Chongqing, China.}
\IEEEauthorblockA{\textsuperscript{4}\,%
Key Laboratory of Perioperative Multi-organ Protection and Intelligent Anesthesia\\
of Chongqing Municipal Health Commission, Chongqing, China}
\IEEEauthorblockA{\textsuperscript{*}\href{mailto:wudi1986@swu.edu.cn}{wudi1986@swu.edu.cn}}
\thanks{This work was supported in part by the New Chongqing Youth Innovation Talent Project and under Grant CSTB2024NSCQ-QCXMX0035, in part by the National Natural Science Foundation of China (62576289, 62176070), and in part by the Key Clinical Research Incubation Project (No.202411TZDA02) of Southwest Hospital, Chongqing, China.}
}

\maketitle

\begin{abstract}
In the intensive care setting, sepsis continues to be a major contributor to patient illness and death; however, its timely detection is hindered by the complex, sparse, and heterogeneous nature of electronic health record ({\ehr}) data.  We propose \tripletgcn, a single-branch graph convolutional model that represents each encounter as patient–feature–value triplets, constructs a bipartite {\ehr} graph, and learns patient embeddings via a Graph Convolutional Network ({\gcn}) followed by a lightweight multilayer perceptron ({\mlp}).  The pipeline applies type-specific preprocessing—median imputation and standardization for numeric variables, effect coding for binary features, and mode imputation with low-dimensional embeddings for rare categorical attributes—and initializes patient nodes with summary statistics, while retaining measurement values on edges to preserve “who measured what and by how much.”  In a retrospective, multi-center Chinese cohort (N = 648; 70/30 train–test split) drawn from three tertiary hospitals, \tripletgcn consistently outperforms strong tabular baselines (KNN, SVM, XGBoost, Random Forest) across discrimination and balanced error metrics, yielding a more favorable sensitivity–specificity trade-off and improved overall utility for early warning.  These findings indicate that encoding {\ehr} as triplets and propagating information over a patient–feature graph produce more informative patient representations than feature-independent models, offering a simple, end-to-end blueprint for deployable sepsis risk stratification.
\end{abstract}


\section{Introduction}
Sepsis is closely associated with substantial increases in both morbidity and mortality among critically ill patients\cite{paper1,paper2}, and timely identification is essential to prevent subsequent organ dysfunction and death.\cite{paper3,paper4}. Driven by the rapid advances in big data\cite{paper5,paper6,paper7,paper8,paper9,paper10,paper11,paper12,paper13,paper14,paper15,paper16,paper17,paper18,paper19,paper20,paper21,paper22,paper23,paper24,paper25,paper26,paper27,paper28}and artificial intelligence technologies\cite{paper29,paper30,paper31,paper32,paper33,paper34,paper35,paper36,paper37,paper38,paper39,paper40,paper41,paper42,paper43,paper44,paper45,paper46,paper47,paper48,paper49,paper50,paper51,paper52,paper53,paper54,paper55},electronic health records in intensive care units contain vast amounts of patient information, which brings unique possibilities as well as huge obstacles to the development of early warning systems for diseases.   Despite the increasing availability of electronic health records ({\ehr}), early warning remains challenging because these data are high-dimensional, sparse, and heterogeneous, evolving across physiological, laboratory, and treatment domains.\cite{paper56}

Conventional machine-learning approaches—such as Ran-dom Forests, XGBoost, Support Vector Machines, Naïve Bayes, and logistic regression—typically treat {\ehr} variables as independent tabular features\cite{paper57,paper58,paper59,paper60,paper61,paper62,paper63}.This independence as-sumption overlooks relational dependencies between patients and clinical attributes, limiting a model’s capacity to capture cross-feature interactions and hindering generalization across diverse patient populations.

In response to these challenges, we introduce a triplet-based graph convolutional framework for sepsis prediction from {\ehr} data\cite{paper64}. Each encounter is transformed into a set of (patient, feature, value) triplets that naturally encode associations between individuals and their observed clinical attributes. We then construct a bipartite graph in which patients and features are distinct node types connected by weighted edges representing measured values; a Graph Convolutional Network ({\gcn}) aggregates signals from each patient’s feature neighbors to learn patient-level embeddings, which are passed to a compact multilayer perceptron ({\mlp})  for risk estimation.

The proposed approach offers a unified and interpretable way to integrate structured {\ehr} information into a graph representation, capturing both intra-patient and inter-patient dependencies while remaining lightweight enough for practical deployment. Conceptually, our model preserves the crucial semantics of 'who measured what and by how much.' The bipartite graph structure directly links a patient (the “who”)to a clinical variable (the “what”), while the edge weight represents the specific measurement (the “by how much”).This preserves the original interaction in a relational form that {\gcn} are designed to exploit; empirically, it provides a principled basis for comparison against strong tabular baselines, including Random Forests, XGBoost, and Support Vector Machines. This work contributes to the growing literature on graph-based clinical prediction by demonstrating that triplet-level {\ehr} graph modeling can enhance the expressive power of {\gcn} for medical risk prediction.

\section{Methodology}
We cast the {\ehr} as a bipartite graph over patients and features, enabling relational reasoning while preserving measurement semantics through edge weights. The overall pipeline comprises: (i) type-specific preprocessing and triplet-to-graph construction; (ii) node initialization for patients and features; (iii) {\gcn} message passing to obtain patient embeddings; and (iv) shallow {\mlp} classification with a focal-loss objective.
\subsection{Problem Setup and Cohort}
Let $\mathcal{P}=\{1,\dots,N\}$ denote the set of patients and $\mathcal{F}=\{1,\dots,M\}$ the set of clinical features extracted from {\ehr}. Each observation is represented as a triplet $(i,f,x_{if})$, where $x_{if}$ is the raw measurement (numeric, binary, or categorical) of feature $f$ for patient $i$. The goal is to learn $h:\mathcal{P}\to[0,1]$ that maps each patient to the probability of sepsis.

Data were collected from three tertiary hospitals in China---Southwest Hospital (Third Military Medical University), Xuanwu Hospital (Capital Medical University), and West China Hospital (Sichuan University)\cite{paper65}. We used a stratified 70/30 train--test split, maintaining an approximately 1:2 positive/negative ratio in both sets\cite{paper66}.

\subsection{Triplet-to-Graph}
 We build a bipartite graph $\mathcal{G}=(\mathcal{V},\mathcal{E},\mathcal{W})$ with node set $\mathcal{V}=\mathcal{V}_P\cup\mathcal{V}_F$, where $\mathcal{V}_P$ are patient nodes and $\mathcal{V}_F$ are feature nodes\cite{paper67,paper68}. For each present (after imputation) measurement, we add an undirected edge $(i,f)\in\mathcal{E}$ between patient $i\in\mathcal{V}_P$ and feature $f\in\mathcal{V}_F$ with scalar weight $w_{if}\in\mathbb{W}$. Self-loops are attached to all nodes for numerical stability\cite{paper69}.

Let $\hat{A}$ be the weighted adjacency with self-loops:
\begin{equation}
\label{eq:adj_def}
\hat{A}_{uv}=
\begin{cases}
w_{if}, & u=i\in \mathcal{V}_{P},~ v=f\in \mathcal{V}_{F},\\
w_{fi}, & u=f\in \mathcal{V}_{F},~ v=i\in \mathcal{V}_{P},\\
1, & u=v,\\
0, & \text{otherwise}.\\
\end{cases}
\end{equation}

We use the standard {\gcn} ``renormalization trick'' with symmetric normalization:
\begin{equation}
\label{eq:renorm}
\tilde{A}=D^{-\frac{1}{2}}\,\hat{A}\,D^{-\frac{1}{2}}
\end{equation}
where the degree matrix $D$ satisfies
\begin{equation}
\label{eq:degree}
D_{uu}=\sum_{v}\hat{A}_{uv}\,
\end{equation}
This graph construction directly models the patient-feature-value triplet: the patient node $i$ (the “who”) is connected to the feature node $f$ (the “what”)by an edge whose scalar weight $w_{if}$ represents the measurement value (the “by how much”).By encoding fine-grained measurements on the edges rather than just the nodes, this design preserves the original clinical semantics and enables meaningful message passing across these specific patient-feature relations.

\subsection{Heterogeneous Feature Handling}
{\ehr} heterogeneity is addressed with type-specific preprocessing applied to the training split and then fixed.

\subsubsection{Numeric features}
Missing values are imputed with the feature-wise median; the result is standardized using training-set statistics:
\begin{equation}
\label{eq:impute}
\tilde{x}_{if}=
\begin{cases}
x_{if}, & \text{if observed},\\
\mathrm{median}_{\mathrm{train}(f)}, & \text{otherwise}.
\end{cases}
\end{equation}
\begin{equation}
\label{eq:standardize}
v_{if}=\frac{\tilde{x}_{if}-\mu_{f}}{\sigma_{f}}\,
\end{equation}

where ${{\mu }_{f}}$, ${{\sigma }_{f}}$ are the training-set mean and standard deviation (with $ {{\sigma }_{f}}\leftarrow 1 $, if degenerate).

\subsubsection{Binary features}
We adopt effect coding without standardization:
\begin{equation}
\label{eq:effectcoding}
v_{if}=
\begin{cases}
+1, & x_{if}=1,\\
-1, & x_{if}=0,\\
0, & \text{missing}.
\end{cases}
\end{equation}

\subsubsection{Categorical features (rare)}
 Categorical attributes are imputed with the training-set mode and embedded via a learnable low-dimensional embedding $e(c)\in\mathbb{R}^{4}$. These embeddings are appended to the patient-node initializer. We additionally record a missingness indicator $m_{if}\in\{0,1\}$ (1 if originally observed, 0 if imputed) to inform edge weighting.

\subsection{Node Initialization}
\subsubsection{Feature nodes}
Each feature $f$ has a trainable embedding $z_f\in\mathbb{R}^{32}$, linearly projected to the {\gcn} hidden size $ d:h_{f}^{\left( 0 \right)}=\psi \left( {{z}_{f}} \right){{\mathbb{R}}^{\text{d}}} $.

\subsubsection{Patient nodes}
 For each patient $i$, we compute summary statistics over incident values $\{v_{if}\}_f$:
\begin{equation}
\label{eq:summarystats}
s_i=\big[\,\mathrm{mean}(v_{if}),~\max(v_{if}),~\min(v_{if}),~\mathrm{var}(v_{if})\,\big]\in\mathbb{R}^{4}
\end{equation}
We then concatenate the categorical embedding(s) $e\left( {{c}_{i}} \right)$ and map to ${{\mathbb{R}}^{d}}$ via a linear projection $\phi$:
\begin{equation}
\label{eq:patientinit}
h_i^{(0)}=\phi\!\left(\,[\,s_i \,\Vert\, e(c_i)\,]\,\right)\in\mathbb{R}^{d} 
\end{equation}

This design supplies patient nodes with coarse, patient-level context while leaving fine-grained measurement information on edges.

\subsection{GCN Model}

\begin{figure*}[!t]
  \centering
  \includegraphics[width=\textwidth]{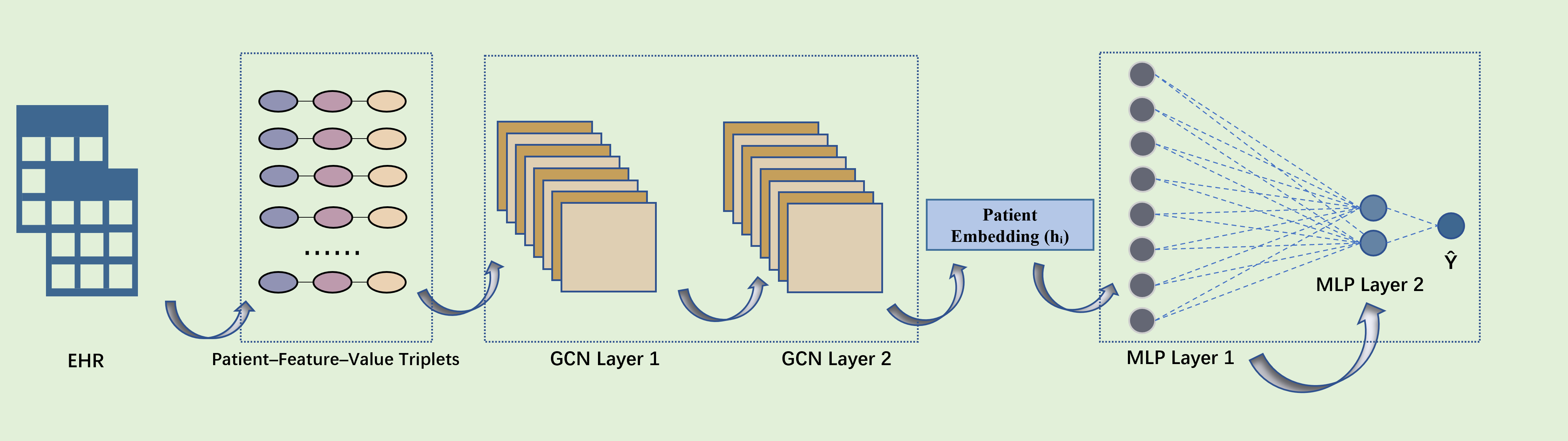}
  \caption{The GCN model framework.}
  \label{fig:fig1}
\end{figure*}

Figure 1 presents the proposed triplet-driven {\gcn}. Patient–feature–value triplets are first encoded, then propagated through a graph convolutional network, and finally classified by a multilayer perceptron to estimate sepsis risk\cite{paper70,paper71,paper72}. The model uses two {\gcn} layers with LeakyReLU activations and dropout\cite{paper73}. The first layer performs message passing with the normalized adjacency $\tilde{A}$; the second incorporates a residual connection to stabilize optimization and retain low-level signals\cite{paper74,paper75,paper76}. For each patient node\cite{paper77}, the embedding is the corresponding row of the final hidden matrix ${{H}^{\left( 2 \right)}}$:
\begin{equation}
\label{eq:gcn1}
H^{(1)} = \sigma\!\left(\tilde{A}\,H^{(0)}\,W^{(0)}\right)
\end{equation}
\begin{equation}
\label{eq:gcn2}
H^{(2)} = \sigma\!\left(\tilde{A}\,H^{(1)}\,W^{(1)}\right) + H^{(1)}
\end{equation}

\subsection{Learning and Inference}
Each patient embedding $h_i=H^{(2)}_i$ is passed to a shallow {\mlp} to obtain a logit $z_i$ and probability $p_i=\sigma(z_i)$. To address the issue of class imbalance while placing greater emphasis on hard-to-classify samples, we adopt the focal loss function\cite{paper78}:
\begin{equation}
\label{eq:focal}
\mathcal{L}=-\frac{1}{N}\sum_{i=1}^{N}\alpha_{t,i}\,\bigl(1-p_{t,i}\bigr)^{\gamma}\,\log p_{t,i}
\end{equation}
where $p_{t,i}$ denotes the predicted probability assigned to the ground-truth class of the $i$-th sample, $\gamma>0$ is the focusing parameter, and $\alpha\in(0,1]$ is a scalar scaling factor.

\section{Results}
Table I presents a detailed comparison of the proposed \tripletgcn model against a set of conventional machine-learning baselines evaluated on the same held-out test set\cite{paper79,paper80}. In terms of overall discriminative capacity, \tripletgcn achieved an AUC of 90.72\%, outperforming the strongest baseline (Random Forest, 89.52\%) by 1.20 percentage points. This improvement in AUC indicates that \tripletgcn is better able to separate septic from non-septic cases across a wide range of decision thresholds. With respect to sensitivity, \tripletgcn attained a value of 60.94\%, corresponding to an absolute increase of 1.56 percentage points over the best-performing baseline (Random Forest, 59.38\%), suggesting enhanced ability to correctly identify true positive cases. Specificity reached 95.42\%, which matches the highest specificity observed among the baselines (KNN, 95.42\%), indicating that the proposed model preserves the strong capacity of existing methods to correctly classify non-septic patients.

Regarding positive predictive performance, \tripletgcn obtained a precision (PPV) of 86.67\%, exceeding the best baseline (KNN, 85.37\%) by 1.30 percentage points, thereby reducing the proportion of false-positive alerts among predicted septic cases. The negative predictive value (NPV) was 83.33\%, representing a 1.14 percentage point gain over the strongest baseline (Random Forest, 82.19\%), which implies a more reliable exclusion of sepsis among patients predicted to be non-septic. Importantly, the F1 score, which jointly reflects precision and recall, reached 71.56\% for \tripletgcn, corresponding to an absolute improvement of 4.30 percentage points relative to the best baseline (Random Forest, 67.26\%). This substantial gain in F1 demonstrates that \tripletgcn achieves a more favorable balance between detecting true sepsis cases and limiting false alarms. Finally, the overall accuracy was 84.10\%, which is 2.05 percentage points higher than the top baseline accuracy (KNN, 82.05\%), further confirming the superior predictive performance of the proposed model.

Taken together, these consistent improvements across multiple complementary metrics indicate that modeling {\ehr} data as patient–feature–value triplets and propagating information over a bipartite graph leads to more expressive and clinically informative patient representations than approaches that treat variables as independent tabular features. By capturing higher-order relationships between patients and their recorded measurements, \tripletgcn enhances discriminative performance while achieving a favorable trade-off between false positives and false negatives in this retrospective cohort.

\begin{table}[htbp]
  \caption{Comparative performance of Triplet--GCN and traditional machine-learning models.}
  \label{tab:perf}
  \centering
  \small
  \setlength{\tabcolsep}{2pt}
  \renewcommand{\arraystretch}{0.95}
  \begin{tabular}{lccccccc}
    \toprule
    \textbf{Model} &
    \shortstack{\textbf{AUC}\\(95\% CI)} &
    \shortstack{\textbf{Sens.}\\(\%)} &
    \shortstack{\textbf{Spec.}\\(\%)} &
    \shortstack{\textbf{PPV}\\(\%)} &
    \shortstack{\textbf{NPV}\\(\%)} &
    \shortstack{\textbf{F1}\\(\%)}  &
    \shortstack{\textbf{Acc.}\\(\%)} \\
    \midrule
    KNN            & 86.66 & 54.69 & 95.42 & 85.37 & 81.17 & 66.67 & 82.05 \\
    SVM            & 87.42 & 53.12 & 92.37 & 77.27 & 80.13 & 62.96 & 79.49 \\
    XGBoost        & 87.79 & 54.69 & 87.02 & 67.31 & 79.72 & 60.34 & 76.41 \\
    Random Forest  & 89.52 & 59.38 & 91.60 & 77.55 & 82.19 & 67.26 & 81.03 \\
    \textbf{Triplet--GCN} & \textbf{90.72} & \textbf{60.94} & \textbf{95.42} & \textbf{86.67} & \textbf{83.33} & \textbf{71.56} & \textbf{84.10} \\
    \bottomrule
  \end{tabular}
\end{table}

\section{Conclusion}
This study introduced \tripletgcn, a lightweight graph-convolutional framework that models {\ehr} as patient–feature–value triplets, propagates information over a patient–feature bipartite graph, and classifies patient embeddings with a compact {\mlp}. By retaining the provenance and magnitude of measurements at the edge level, the method unifies heterogeneous variables in a single relational representation while remaining interpretable and practical for deployment\cite{paper81,paper82}. 

On a multi-institutional Chinese cohort, \tripletgcn consistently outperformed strong tabular baselines across discrimination and balanced-error metrics, directly addressing the question of whether graph-based modeling of triplet-structured {\ehr} yields tangible benefits for clinical risk stratification. The findings indicate that encoding relations at the edge level improves {\gcn}-derived patient representations and enhances robustness to sparsity and heterogeneity in real-world data. 

Looking ahead, we will pursue external and prospective validation and incorporate temporal dynamics to better capture disease trajectories, while preserving the simplicity and interpretability of the approach. These steps are natural extensions toward evaluating generalizability, operational readiness, and longitudinal risk modeling in clinical settings.

\bibliographystyle{IEEEtran}
\bibliography{ISCEIC} 
\end{document}